\pdfoutput=1

\documentclass[11pt]{article}

\usepackage[]{EMNLP2022}

\usepackage{times}
\usepackage{latexsym}

\usepackage[T1]{fontenc}

\usepackage[utf8]{inputenc}

\usepackage{microtype}

\usepackage{inconsolata}
\usepackage{amsmath}
\usepackage{multirow}
\usepackage{booktabs}
\usepackage{graphicx}

\title{Improved Data Augmentation for Translation Suggestion}

  \author{Hongxiao Zhang\textsuperscript{1}, Siyu Lai\textsuperscript{1}, 
  Songming Zhang\textsuperscript{1}, 
  Hui Huang\textsuperscript{2},
  \textbf{Yufeng Chen}\textsuperscript{1}\thanks{\ \ Yufeng Chen is the corresponding author.}\\ 
  \textbf{Jinan Xu\textsuperscript{1}} and \textbf{Jian Liu\textsuperscript{1}}\\
\textsuperscript{1}Beijing Jiaotong University, Beijing, China \\
\textsuperscript{2}Harbin Institute of Technology, Harbin, China \\
\texttt{\{hongxiaozhang,siyulai,smzhang22,chenyf,jaxu,jianliu\}@bjtu.edu.cn}, \\
\texttt{huanghui\_hit@126.com} \\}

\begin{document}
\maketitle
\begin{abstract}
Translation suggestion (TS) models are used to automatically provide alternative suggestions for incorrect spans in sentences generated by machine translation. This paper introduces the system used in our submission to the WMT'22 Translation Suggestion shared task. Our system is based on the ensemble of different translation architectures, including Transformer, SA-Transformer, and DynamicConv. We use three strategies to construct synthetic data from parallel corpora to compensate for the lack of supervised data. In addition, we introduce a multi-phase pre-training strategy, adding an additional pre-training phase with in-domain data. 
We rank second and third on the English-German and English-Chinese bidirectional tasks, respectively.
\end{abstract}

\section{Introduction} \label{Introduction}

Translation suggestion (TS) is a scheme to simplify Post-editing (PE) by automatically providing alternative suggestions for incorrect spans in machine translation outputs. \citet{yang2021wets} formally define TS and build a high-quality dataset with human annotation, establishing a benchmark for TS. Based on the machine translation framework, the TS system takes the spliced source sentence $\mathbf{x}$ and the translation sentence $\mathbf {\Tilde{m}}$ as the input, where the incorrect span of $\mathbf {\Tilde{m}}$ is masked, and its output is the correct alternative $\mathbf y$ of the incorrect span. The TS task is still in the primary research stage, to spur the research on this task, WMT released the translation suggestion shared task.

This WMT'22 shared task consists of two subtasks: Naive Translation Suggestion and Translation Suggestion with Hints. We participate in the former, which publishes the bidirectional translation suggestion task for two language pairs, English-Chinese and English-German, and we participate in all language pairs.

Our TS systems are built based on several machine translation models, including Transformer \citep{vaswani2017attention}, SA-Transformer \citep{yang2021wets}, and DynamicConv \citep{wu2018pay}. To make up for the lack of training data, we use parallel corpora to construct synthetic data, based on three strategies. Firstly, we randomly sample a sub-segment in each target sentence of the golden parallel data, mask the sampled sub-segment to simulate an incorrect span, and use the sub-segment as an alternative suggestion. Secondly, the same strategy as above is used for pseudo-parallel data with the target side substituted by machine translation results. Finally, we use a quality estimation (QE) model \citep{zheng2021self} to estimate the translation quality of words in translation output sentence and select the span with low confidence for masking, and then, we utilize an alignment tool to find the sub-segment corresponding to the span in the reference sentence and use it as the alternative suggestion for the span.

Considering that there is a domain difference between the synthetic corpus and the human-annotated corpus, we add an additional pre-training phase. Specifically, we train a discriminator and use it to filter sentences from the synthetic corpus that are close to the golden corpus, which we deem as in-domain data. After pre-training with large-scale synthetic data, we perform an additional pre-training with in-domain data, thereby reducing the domain gap. We will describe our system in detail in Section \ref{Method}.

\section{Related Work} \label{Related_Work}

The translation suggestion (TS) task is an important part of post-editing (PE), which combines machine translation (MT) and human translation (HT), and improves the quality of translation by correcting incorrect spans in machine translation outputs by human translators. To simplify PE, some early scholars have studied translation prediction (\citet{green2014human}, \citet{knowles2016neural}), which provides predictions for the next word (or phrase) when given a prefix. And some scholars have also studied prediction with the hints of translators \citep{huang2015new}.

In recent years, some scholars have devoted themselves to researching methods to provide suggestions to human translators. \citet{santy2019inmt} present a proof-of-concept interactive translation system that provides human translators with instant hints and suggestions. \citet{lee2021intellicat} utilize two quality estimation models and a translation suggestion model to provide alternatives for specific words or phrases for correction. \citet{yang2021wets} propose a transformer model based on segment-aware self-attention, provide strategies for constructing synthetic corpora, and released the human-annotated golden corpus of TS, which became a benchmark for TS tasks.

\section{Method} \label{Method}

In this section, we describe the translation suggestion system, followed by our strategies for building synthetic corpora, and finally the details of the additional pre-training phase.

\subsection{Translation Suggestion System}

As defined by \citet{yang2021wets}, given the source sentence $\mathbf{x}$, its translation sentence $\mathbf{m}$, the incorrect span $\mathbf{w}$ in $\mathbf{m}$, and its corresponding correct translation $\mathbf{y}$, the translation suggestion task first masks the incorrect span $\mathbf{w}$ in $\mathbf{m}$ to get $\mathbf{m^{-w}}$, and then maximizes the following conditional probabilities:
\begin{equation}
    p(\mathbf{y} | \mathbf{x},\mathbf{m^{-w}}; \boldsymbol{\theta})
\end{equation}
\noindent where $\boldsymbol{\theta}$ is the parameters of the model.

The construction of the TS system is based on common machine translation models. We introduce the models used in our TS system below:

\begin{itemize}
    \item \textbf{Transformer-base \citep{vaswani2017attention}.} The naive transformer model. The encoding and decoding layers are both set to 6, the word embedding size is set to 512, and the attention head is set to 8.
    \item \textbf{Transformer-big \citep{vaswani2017attention}.} The widened transformer model. The encoding and decoding layers are both set to 6, the word embedding size is set to 1024, and the attention head is set to 16.
    \item \textbf{SA-Transformer \citep{yang2021wets}.} The segment-aware transformer model, which replaces the self-attention of the naive transformer with the segment-aware self-attention, further injects segment information into the self-attention, so that it behaves differently according to the segment information of the token. Its parameter settings are the same as those of Transformer-base.
    \item \textbf{DynamicConv \citep{wu2018pay}.} The dynamic convolution model that predicts a different convolution kernel at every time-step. We set both encoding GLU and decoding GLU to 1 in the experiment.
\end{itemize}

\subsection{Build Synthetic Corpora}
\label{build}
Since there are few golden corpora available for training, it is necessary to build a synthetic corpus to make up for the lack of data. We build synthetic data through the following three strategies and use the mixed data for model pre-training.

\subsubsection{Building on Golden Parallel Data}
\label{golden}

Following the method of \citet{yang2021wets}, we construct synthetic data on the golden parallel corpus. Given a sentence pair $\mathbf x = \{x_1, x_2, \ldots, x_n \}$ and $\mathbf r = \{ r_1, r_2, \ldots, r_m \}$ from the golden parallel corpus, we randomly sample a sub-segment $\mathbf w = \{r_i, r_{i+1}, \ldots, r_j\}$ of $\mathbf r$, we mask the sub-segment in sentence $\mathbf r$ to get $\mathbf{r^{-w}} = \{r_1, r_2, \ldots, r_{i-1},\mathrm{[MASK]},r_{j+1}, \ldots, r_m\}$, and use $\mathbf w$ as an alternative suggestion. We perform statistics on the length of golden data to determine the length of masked spans, which is more in line with the golden distribution.

\subsubsection{Building on Pseudo Parallel Data}

The prediction of alternative suggestions requires the translation context, which cannot be provided by the golden parallel corpus. Therefore, we still follow \citet{yang2021wets} and use the same way as described in Section \ref{golden} to construct synthetic data on the pseudo-parallel corpora consisting of source sentences and machine translation output sentences. 

\subsubsection{Building with Quality Estimation}

\begin{figure*}
    \centering
    \includegraphics[scale=0.45]{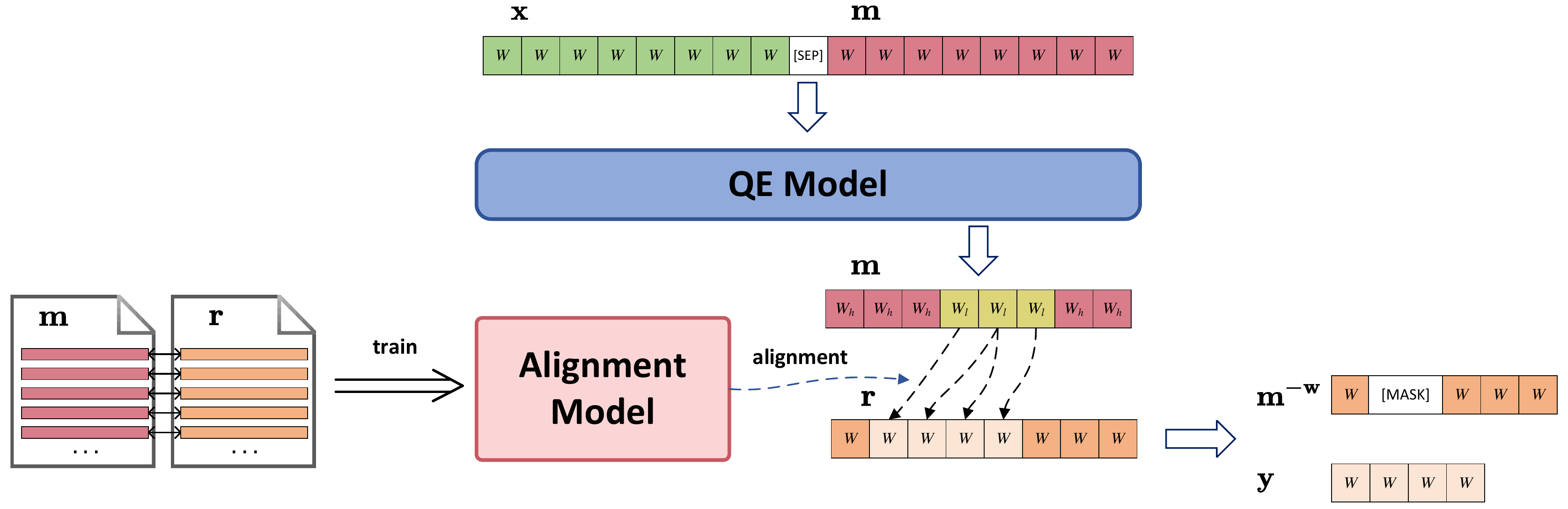}
    \caption{Schematic diagram of building synthetic corpora with quality estimation. $\mathbf x$ is the source sentence, $\mathbf m$ is the machine translation sentence, $\mathbf r$ is the reference sentence, and $W_h$ and $W_l$ represent words with high and low confidence, respectively.}
    \label{fig:qe}
\end{figure*}

The TS task is to predict the correct alternative proposal given the translation context. However, when sampling on the golden parallel corpus, the context does not match the translation output, and when sampling on the pseudo-parallel corpus, the alternative suggestions may be incorrect. Therefore, the above two construction strategies are not optimal.

We explore a method that is closer to the real scenarios, as shown in Figure~\ref{fig:qe}. First, the word-level translation quality estimation (QE) model is used to estimate the confidence of the words in the translation sentence, and the continuous span with low confidence (that is, poor translation) is selected. Then, the translation sentence is aligned with the reference sentence through the alignment model, and the sub-segment corresponding to the span in the reference is selected as an alternative suggestion.

More specifically, we use a masked language model as our QE model, following the method of \citet{zheng2021self}. To train the QE model, we splice the source sentence $\mathbf x_i$ and the reference sentence $\mathbf r_i$ of the golden parallel corpus, where some words in $\mathbf r_i$ are masked to get $\mathbf r^{-w}_i$, and the QE model is optimized to minimize the following loss function:

\begin{equation}
    \mathcal{L} = -\sum_{i=1}^N \log p(\mathbf{r}^w_i | \mathbf{x}_i,\mathbf{r}^{-w}_i; \boldsymbol{\theta})
\end{equation}

\noindent where $N$ is the number of golden parallel sentences, $\mathbf{r}^w_i$ is the masked part of the reference sentence and $\boldsymbol{\theta}$ is the model parameter.

During inference, the source and translation sentences of the pseudo-parallel corpus are spliced and fed into the QE model. The model scores the word of the translation sentence according to the recovery probability of it after being masked, and words with lower scores are considered poor translations.

After that, we train a word alignment model \citep{Lai2022cross} using the translated sentences and reference sentences. To ensure high alignment quality, we filter out sentences with lengths less than 5 and greater than 100 and randomly sample 5M sentence pairs for training. We use the trained alignment model to align the machine translation sentence and the reference sentence. The sub-segment in the reference that aligns with the poorly translated words described above is selected as alternative suggestion.


\subsection{Additional Pre-Training Phase with In-Domain Data}

The sources of data used to construct large-scale synthetic corpus and human-annotated golden corpus are domain different. To bridge this difference, we introduce an additional pre-training stage. We filter data similar to the golden corpus as in-domain data, which are used as pre-training for the next phase after pre-training model with a large-scale synthetic corpus.

In particular, we use BERT \citep{devlin2019bert} to construct a discriminator to identify in-domain data. The discriminator consists of a binary classifier trained to distinguish between in-domain and out-of-domain sentences. The source sentences from the golden corpus as positive examples and source sentences from the synthetic corpus as negative examples are used to train this discriminator. We upsample the golden corpus to 10 times, and randomly subsample the same amount of sentences from the synthetic corpus. For each input source sentence, the discriminator predicts the probability that the sentence is in-domain. Sentences with probabilities greater than a certain threshold are discriminated as in-domain sentences.

After the above two phases of pre-training, we use the human-annotated golden corpus for fine-tuning and test the final model.

\section{Experiments and Results}

\subsection{Setup}

\begin{table}[]
    \centering
    \begin{tabular}{lc c c c}
    \midrule[1pt]
        \textbf{Corpus} & \textbf{golden}  & \textbf{pseudo} & \textbf{with QE} \\
    \hline
       LS en$\Leftrightarrow$de & 9.8M & 9.8M & 4.7M \\
       LS en$\Leftrightarrow$zh & 20M & 20M & – \\
       IND en$\Rightarrow$de & 0.8M  &  0.8M  & 0.4M \\
       IND de$\Rightarrow$en & 0.7M & 0.7M & 0.3M \\
     \midrule[1pt]
    \end{tabular}
    \caption{Statistics of constructed synthetic data in our experiments, where LS stands for large-scale data and IND stands for in-domain data.}
    \label{tab:data}
\end{table}

\begin{table}[]
    \centering
    \begin{tabular}{c|c c c c}
    \midrule[1pt]
    \multirow{2}*{System} & \multicolumn{4}{c}{Translation direction} \\
     & zh-en & en-zh & de-en & en-de\\
    \hline
       Baseline & 25.51 & \textbf{36.28} & 31.20 & 29.48 \\
       Ours & \textbf{28.56} & 33.33 & \textbf{36.30} & \textbf{42.61} \\
    \midrule[1pt]
    \end{tabular}
    \caption{BLEU scores on the WMT 2022 TS test set.}
    \label{tab:result}
\end{table}

We have submitted English-Chinese (en-zh) and English-German (en-de) bidirectional translation suggestion tasks. We mix en-zh data from  WMT'19 and WikiMatrix, and en-de data from WMT'14 and WikiMatrix, respectively, to construct a synthetic dataset. We follow \citet{yang2021wets} to preprocess the data, and mix the data constructed by the three strategies described in Section \ref{build}  as our large-scale synthetic data. The statistics of the constructed large-scale (LS) synthetic data and in-domain (IND) synthetic data are shown in Table \ref{tab:data}. Note that for the experiments in the en-zh translation direction, we do not apply the construction strategy with QE and the pre-training phase with in-domain data. All our models are implemented based on Fairseq \citep{ott2019fairseq}. We use the same data on each model for two phases of pre-training and fine-tuning.

\subsection{Results}

We report the results of our method on the development and test set of the translation suggestion task of WMT'22. SacreBLEU\footnote{\url{https://github.com/mjpost/sacrebleu}} is used to compute the BLEU score as quality estimates relative to a human reference. We report the experimental results of our system and the baseline system \citep{yang2021wets} on the test set in Table \ref{tab:result}, and for the baseline system, we directly use their experimental results.

\begin{table}[]
    \centering
    \begin{tabular}{l|c}
    \midrule[1pt]
        \textbf{System} & \textbf{BLEU} \\
        \hline
        Do nothing & 18.24 \\
        \ + on golden and pseudo corpus & 26.91 \\
        \ + with quality estimation & 30.72 \\
        \ + IND pre-training phase & 32.95 \\
    \midrule[1pt]
    \end{tabular}
    \caption{BLEU scores on the English-German development set for systems based on the SA-Transformer model under different strategies.}
    \label{tab:strategy}
\end{table}

As can be seen from Table \ref{tab:result}, our system beats the baseline system in three translation directions, especially in the en-de direction, where our system surpasses the baseline by 13.13 BLEU. 

\begin{table}[]
    \centering
    \begin{tabular}{c|c}
    \midrule[1pt]
        \textbf{Model} & \textbf{BLEU} \\
    \hline
        Transformer-base (A) &  32.92 \\
        Transformer-big (B)  &  34.73 \\
        SA-Transformer (C)  &  32.95 \\
        DynamicConv (D)  &  34.03 \\
        Ensemble (A + B + C + D) & \textbf{35.81} \\
        
    \midrule[1pt]
    \end{tabular}
    \caption{BLEU scores on the development set for systems under different models in the English-German direction.}
    \label{tab:result_dev}
\end{table}

We also report the results of the system on the development set of English-German translation directions to analyze the effectiveness of different models and strategies. In Table \ref{tab:strategy}, we show the results of the system based on the SA-Transformer model under different strategies. ``Do nothing'' means we only train with the provided training set. It can be seen that the strategy of constructing synthetic data with quality estimation (QE) and the additional pre-training with the in-domain (IND) data stage  can bring about a great improvement. 

In Table \ref{tab:result_dev}, we present the results of systems based on different models and the model ensemble. The ensemble model brings obvious improvement and achieves the best results. 

\section{Conclusion}

 We describe our contribution to the Translation Suggestion Shared Task of WMT'22. We propose a strategy to construct synthetic data with the quality estimation model to mask the constructed data closer to the real scenarios. Furthermore, we introduce an additional phase of pre-training with in-domain data to reduce the gap between synthetic corpus and golden corpus. Experimental results demonstrate the effectiveness of our strategy. Considering the heavy labor of annotating TS data, we think data augmentation is the most important strategy that should be addressed. In the future, we will put more effort into the data generation method, to make the most of openly-accessible parallel data.

\section*{Limitations}

The strategy of constructing synthetic data based on quality estimation proposed in this paper can automatically sample the incorrectly translated spans in the translations, and find the correct alternative suggestions through the alignment. It is a solution that conforms to the real scenarios, and the experimental results have also proved that it is effective. However, in our experiments, we find that the quality estimation and alignment phases require a large additional time overhead, and we hope to explore more efficient solutions in future research.


\section*{Acknowledgements}
The research work descried in this paper has been supported by the National Key R\&D Program of China (2020AAA0108001) and the National Nature Science Foundation of China (No. 61976016, 61976015, and 61876198). The authors also would like to thank the WMT'22 shared task organizers for organizing this competition and for providing open source code and models.

\bibliography{emnlp2022}
\bibliographystyle{acl_natbib}




\end{document}